\documentclass[lettersize,journal]{IEEEtran}
\usepackage{amsmath,amsfonts}
\usepackage{algorithmic}
\usepackage{array}
\usepackage[caption=false,font=normalsize,labelfont=sf,textfont=sf]{subfig}
\usepackage{textcomp}
\usepackage{stfloats}
\usepackage{url}
\usepackage{verbatim}
\usepackage{verbatim}
\usepackage{multirow}
\usepackage{graphicx}
\usepackage{cite}
\usepackage{soul}
\usepackage{color, xcolor} 
\usepackage{booktabs}
\usepackage{pdflscape}
\usepackage{amsmath}
\usepackage{array} 
\usepackage{multirow}
\usepackage{graphicx}
\def\BibTeX{{\rm B\kern-.05em{\sc i\kern-.025em b}\kern-.08em
    T\kern-.1667em\lower.7ex\hbox{E}\kern-.125emX}}
\usepackage{balance}
\begin{document}
\title{Efficient Prompt Tuning of Large Vision-Language Model for Fine-Grained Ship Classification}
\author{
\IEEEauthorblockN{Long Lan$^{\dagger }$, \IEEEmembership{Member, IEEE,} Fengxiang Wang$^{\dagger }$,  Xiangtao Zheng, \IEEEmembership{Senior Member, IEEE,}\\ Zengmao Wang, \IEEEmembership{Member, IEEE} and Xinwang Liu, \IEEEmembership{Senior Member, IEEE}}
\thanks{
Fengxiang Wang and Long Lan are with the College of Computer Science and Technology, National University of Defense Technology, Changsha 410073, China. E-mail: \{wfx23, long.lan\}@nudt.edu.cn.}
\thanks{
Xiangtao Zheng is with the College of Physics and Information Engineering, Fuzhou University, Fuzhou 350108, China. E-mail: xiangtaoz@gmail.com.}
\thanks{
Zengmao Wang is with School of Computer Science, Wuhan University, Wuhan 430000, China. E-mail: wangzengmao@whu.edu.cn.}
\thanks{
 Xinwang Liu is with the College of Computer, National University of Defense Technology, Changsha 410073, China. E-mail: xinwangliu@nudt.edu.cn.}

\thanks{ Corresponding Author: Fengxiang Wang.}
 \thanks{$^{\dagger }$ Equal contribution}
 }

\markboth{Journal of \LaTeX\ Class Files,~Vol.~18, No.~9, September~2023}%
{How to Use the IEEEtran \LaTeX \ Templates}

\maketitle

\begin{abstract}
Fine-grained ship classification in remote sensing (RS-FGSC) poses a significant challenge due to the high similarity between classes and the limited availability of labeled data, limiting the effectiveness of traditional supervised classification methods. Recent advancements in large pre-trained Vision-Language Models (VLMs) have demonstrated impressive capabilities in few-shot or zero-shot learning, particularly in understanding image content. This study delves into harnessing the potential of VLMs to enhance classification accuracy for unseen ship categories, which holds considerable significance in scenarios with restricted data due to cost or privacy constraints. Directly fine-tuning VLMs for RS-FGSC often encounters the challenge of overfitting the seen classes, resulting in suboptimal generalization to unseen classes, which highlights the difficulty in differentiating complex backgrounds and capturing distinct ship features. To address these issues, we introduce a novel prompt tuning technique that employs a hierarchical, multi-granularity prompt design. Our approach integrates remote sensing ship priors through bias terms, learned from a small trainable network. This strategy enhances the model's generalization capabilities while improving its ability to discern intricate backgrounds and learn discriminative ship features. Furthermore, we contribute to the field by introducing a comprehensive dataset, FGSCM-52, significantly expanding existing datasets with more extensive data and detailed annotations for less common ship classes. Extensive experimental evaluations demonstrate the superiority of our proposed method over current state-of-the-art techniques. The source code will be made publicly available.
\end{abstract}


\section{Introduction} 
\IEEEPARstart{A}{s}  maritime activities continue to grow, the demand for accurate Remote Sensing Fine-Grained Ship Classification (RS-FGSC) has surged, necessitating advanced methodologies to cope with the intricacies of ship identification \cite{r2,r4}. Deep neural networks have marked significant progress in RS-FGSC, with numerous studies showcasing their efficacy \cite{r5,r6,r7,r8,r9,r10,zhang2022b,zhang2022a,zhang2023a}.
Despite these advancements, RS-FGSC remains a formidable challenge, particularly when dealing with unseen data and the scarcity of labeled samples. The acquisition of data for certain ship categories is often hindered by practical constraints, such as cost and privacy concerns.
To tackle this, few-shot learning strategies have been proposed to enhance classification performance with limited training data \cite{r11,r12,r13,metaprompt}. However, these methods, which are typically trained on a small number of labeled images and evaluated on datasets with the same categories, may not adequately address the complexities of real-world applications.
In response to these challenges, particularly in the context of practical remote sensing, there is a pressing need to improve the model's ability to generalize to previously unseen ship classes. This is known as the base-to-new generalized classification task \cite{coop,clip}, which assumes that new classes differ from the base classes, thereby testing the model's capacity for generalization across diverse categories.

\begin{figure}[!t]
    \centering 	
    \subfloat[]{\includegraphics[width=0.22\textwidth,height=20em]{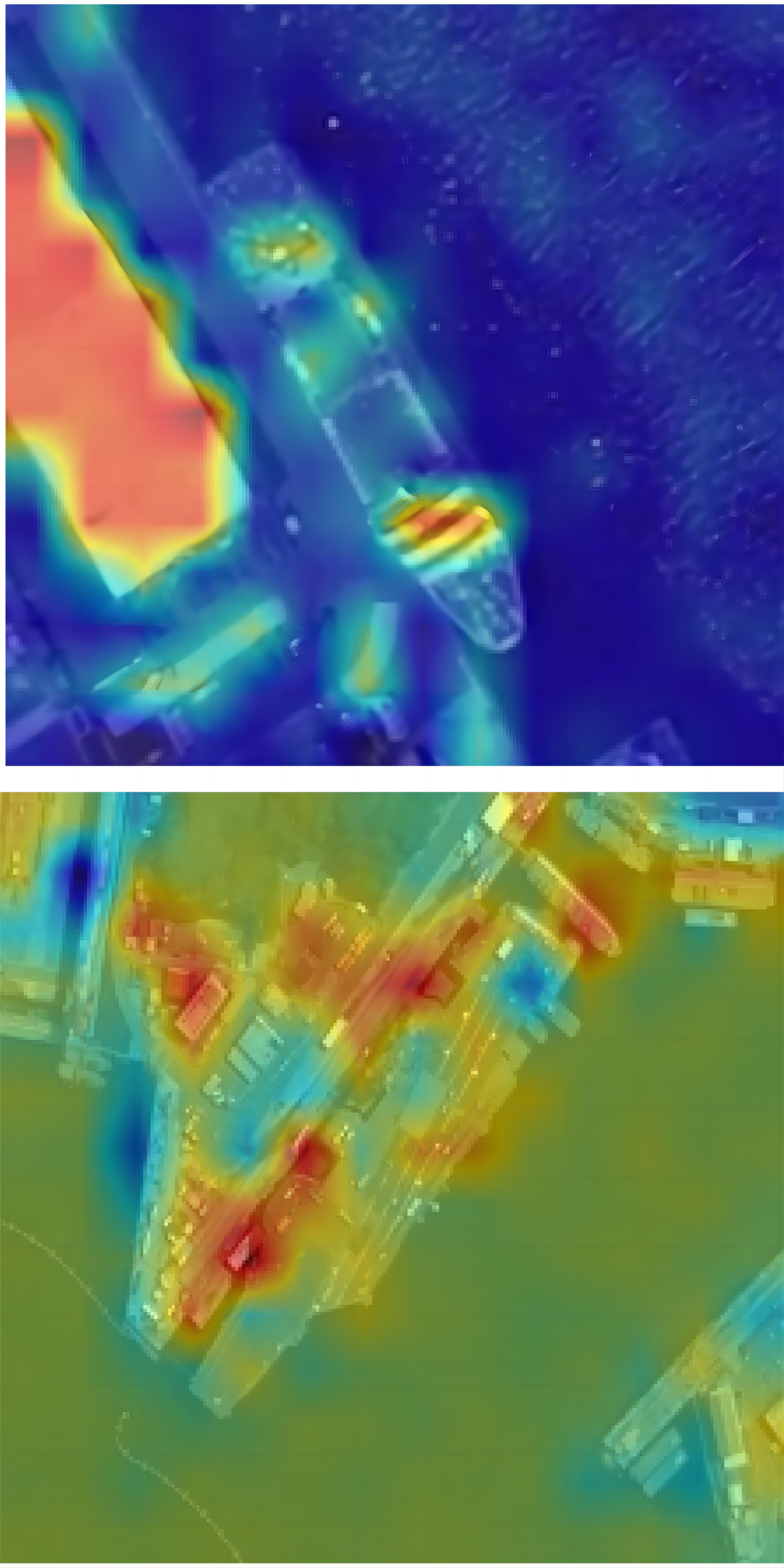}}	
    \hspace{0.5em}
    \subfloat[]{\includegraphics[width=0.22\textwidth,height=20em]{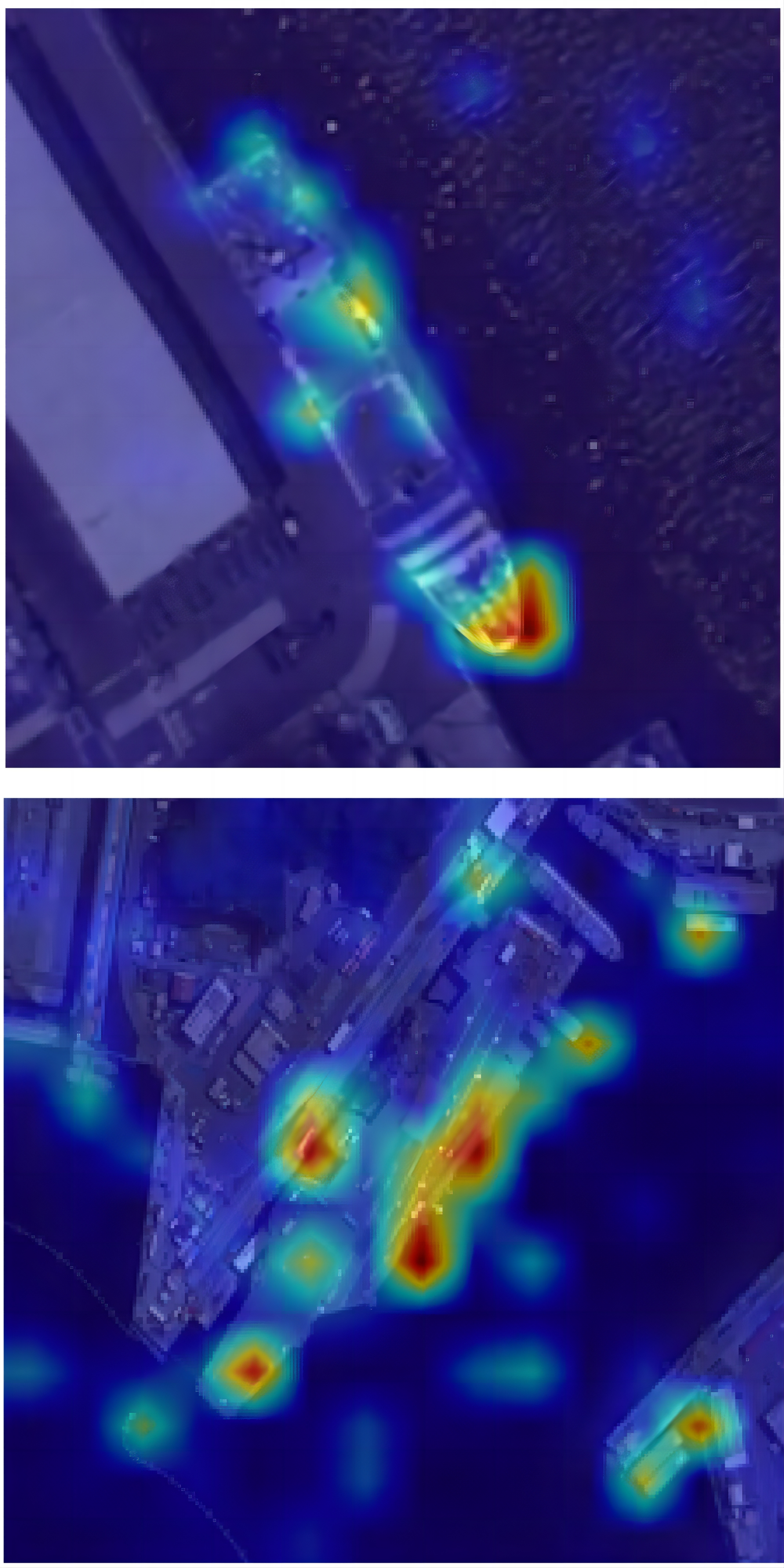}}
    \caption{Visualization of the attention maps of (a) the baseline CLIP~\cite{clip} and (b) the proposed method. Our method concentrates on ship targets while the CLIP model attends to the background excessively.}
    \label{fig1}
\end{figure}

Large pre-trained Vision-Language Models (VLMs) have recently emerged as a powerful tool for vision tasks, with models such as Contrastive Language-Image Pretrainin (CLIP) showcasing their potential by transforming classification into an image-text matching problem through self-supervised learning \cite{clip}. These models, trained on vast datasets of images and text, have achieved performance levels comparable to fully supervised techniques \cite{vlm,RR2,RR3,RR4,RR5,RR6,RR7,RR8,RR9,RR10,clip,RR12,RR13,RR14,shuyan1,shuyan2}. Despite their promise, deploying CLIP-based methods for RS-FGSC often results in overfitting to base classes, hindering generalization to novel classes. 

Prompt-tuning methods, such as CoOp \cite{coop}, which fine-tune CLIP using learnable textual parameters (prompts), have shown significant performance improvements in downstream tasks. This approach represents a shift towards a more dynamic and adaptable prompt approach, enhancing VLMs' effectiveness in base-to-new generalization. However, these methods still struggle in the overfitting challenge, particularly in RS-FGSC, since the model's feature representation is biased towards general natural images due to its pre-training stage. As illustrated in Fig. \ref{fig1}, the CLIP model for RS-FGSC attends to the background excessively, often overlooking critical ship features.

To address these issues, we introduce a novel hierarchical, multi-granularity prompt design that integrates three levels of ship category description to provide a comprehensive prompt to the CLIP model. Moreover, to bridge the domain gap between remote sensing images and the natural images used for CLIP training, we propose integrating remote sensing ship priors into the model. This is achieved by refining textual prompts and image features through bias terms learned from small, specialized networks. We also learn task-specific prompts to effectively narrow the gap between base and new classes, enabling fast adaptation to novel classes. Our approach combines multi-granularity textual prompts with domain-specific prior knowledge to reduce the domain gap and learn generalizable features for efficient base-to-new ship classification. Furthermore, we introduce the FGSCM-52 dataset, an enhanced version of FGSCR-42, featuring an additional ten categories and an increased number of samples for existing categories. It can serve as a testbed for RS-FGSC study, aiding in addressing the challenges of fine-grained ship classification effectively.

The main contributions of our research are:
\begin{itemize}
\item We introduce the base-to-new generalization task in RS-FGSC, identify its challenges, and propose utilizing VLMs to enhance model adaptability. We also establish a comprehensive dataset to advance research in this field by extending the existing dataset with more data and annotations for rare classes.
\item We present a novel prompt tuning method for the base-to-new generalization task in RS-FGSC, leveraging a hierarchical, multi-granularity prompt design and incorporating remote sensing ship priors to learn a generalizable feature representation.
\item Our extensive experiments across various datasets demonstrate that our proposed method significantly outperforms state-of-the-art methods.
\end{itemize}

\begin{figure*}[t]
\centering
\includegraphics[width=1\textwidth]{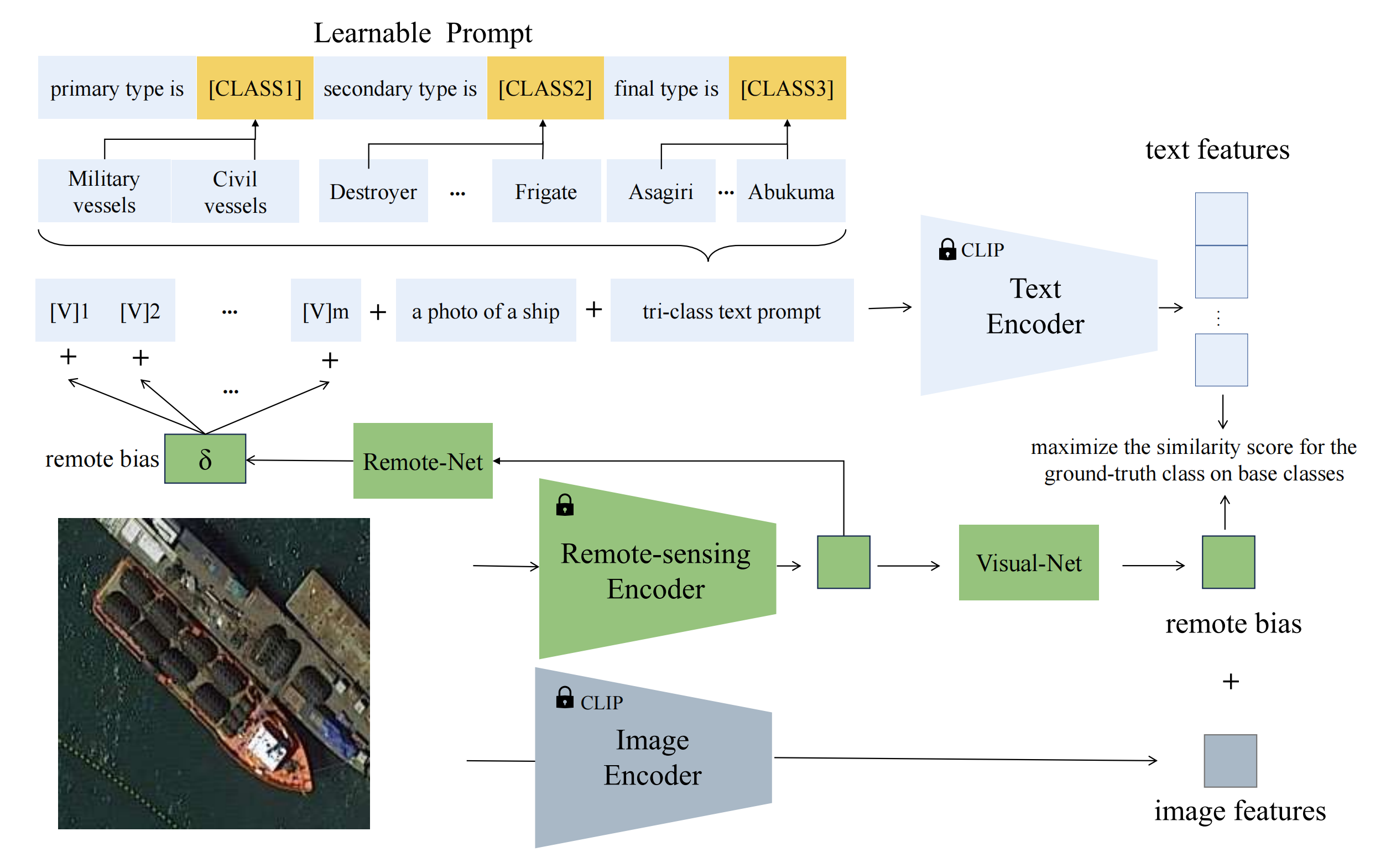}
\caption{Overview of the proposed method. 
The main idea is that the model leverages a hierarchical, multi-granularity prompt design and incorporates remote sensing ship priors to learn generalizable feature representation.}
\label{fig3}
\end{figure*}

\section{Related Work}
\subsection{Fine-grained ship classification}
The evolution of RS-FGSC has seen a transition from basic ship detection to more nuanced categorization. Early research primarily concentrated on distinguishing ships from non-ship objects and broad classification into categories such as warships and civilian vessels \cite{r2,r3}. The advent of large-scale datasets like FGSC-23 \cite{fgsc23} and FGSC-42 \cite{fgsc42} has significantly propelled the study of fine-grained classification, providing a rich resource for researchers to tackle the key challenges of RS-FGSC, including inter-class similarity and class imbalance. Zhang et al. \cite{r4} introduced a technique for extracting both coarse and fine-grained ship features, enhancing the model's ability to discern subtle differences between ship classes. Zhao et al. \cite{r5} proposed a feature balancing strategy that leverages both super-resolution and low-resolution images, particularly advantageous for low-resolution classification tasks. Chen \cite{r6} utilized hierarchical and exclusion graphs to improve the understanding of label hierarchies, while an attention model \cite{r7} was developed to enhance ship detection in dense and complex scenes. Other approaches such as the use of style transfer networks \cite{r8} to simulate diverse ship samples have been employed to address the issue of limited training data, effectively increasing the variety and volume of training data. Contrastive learning strategies \cite{r9} and explainable attention networks \cite{r10} have also been proposed to improve classification accuracy and model interpretability.

Despite these advancements, the acquisition of ship images remains challenging, leading to the development of few-shot learning methods. A metric-based few-shot method \cite{r11} dynamically generates category prototypes by computing nearest neighbor values, while a foreground-aware FRN \cite{r12} reconstructs query features from support features using ridge regression and predicts category distributions by comparing weighted distances with foreground weights. Another study \cite{r13} proposed a generalized ridge-regression-based channel feature map weighted reconstruction network to address the few-shot classification problem. However, current few-shot learning and fully supervised methods still struggle with the base-to-new generalization challenge in RS-FGSC. The existing datasets, while valuable, are limited in scope and quality, necessitating the development of more comprehensive and diverse datasets to support the advancement of RS-FGSC research.

\subsection{Prompt tuning of large vision-language model}
VLMs have revolutionized the field of computer vision by learning generic cross-modal representations from vast image-text corpora, enabling them to be fine-tuned for a variety of downstream visual-linguistic tasks \cite{vlm}. VLMs are categorized into three architectural types: fusion encoders (both single-stream \cite{RR8,RR10,RR13} and dual-stream \cite{RR2,RR3,RR4}), dual encoders \cite{RR7,RR9,clip}, and hybrid models \cite{RR6,RR12}. Among these, the dual encoder model, CLIP, stands out for its use of approximately 400 million image-text pairs for unsupervised learning \cite{clip}.

To improve CLIP's performance in base-to-new generalization tasks, where the model must generalize to unseen categories, methods such as adapters \cite{RR43} and prompt tuning \cite{prompttuning} have emerged as effective strategies. Adapters \cite{RR43, RR44,RR45} introduce a residual feature blending module for efficient transfer learning of VLMs. Prompt tuning, on the other hand, automates the prompt design process, utilizing the labeled data available in downstream tasks. For example, CoOp \cite{coop} is a notable approach that replaces hand-crafted prompts with a set of learnable prompts inferred from labeled few-shot samples. However, CoOp's learnable prompts are unique and fixed for each task's images, which can limit its flexibility. CoCoOp \cite{cocoop} generates an image-conditional context for each image and combines it with a textual-conditional context for prompt tuning. ProGrad \cite{prograd} optimizes only the prompts whose gradients align with the general direction of the task, while KgCoOp \cite{kgcoop} enhances the applicability of learnable prompts by minimizing the discrepancy between the textual embeddings generated by learned prompts and those in CLIP.

Our work builds upon CoOp and its successors, leveraging the expressive power of VLMs for the base-to-new generalization task. However, in the RS-FGSC domain, CoOp-based methods often struggle with overfitting, particularly to complex backgrounds in base class data, which compromises their ability to generalize to new classes. This limitation underscores the need for a more robust prompt tuning approach that can effectively adapt to the unique challenges of RS-FGSC.
Our proposed remote sensing prompt-tuning method relies on VLMs, keeping their text and image encoders fixed for feature extraction.
The method involves two stages: prompt training and the subsequent testing stage.
In the training stage, our method employs the hierarchical, multi-granularity prompt design and remote sensing ship priors for training on base class data.
During the testing phase, we leverage our prompt design to enable predictions for new class data in the base-to-new generalization task.

\section{Method}
Our proposed remote sensing prompt-tuning method relies on VLMs, keeping their text and image encoders fixed for feature extraction.
The method involves two stages: prompt training and the subsequent testing stage.
In the training stage, our method employs the hierarchical, multi-granularity prompt design and remote sensing ship priors for training on base class data.
During the testing phase, we leverage our prompt design to enable predictions for new class data in the base-to-new generalization task.

\subsection{Remote sensing ship priors}
The CLIP model \cite{clip}, distinguished by its extensive training on 400 million image-text pairs, has demonstrated exceptional generalization ability for image classification across a wide array of new class data. This model, featuring both visual and text encoders, is a cornerstone of our approach.
In the prompt training stage, we initially employ and freeze the two computationally intensive yet powerful encoders from CLIP. Subsequently, we design two corresponding lightweight encoders to augment their capabilities. An architectural overview is provided in Fig. \ref{fig3} for clarity.

\begin{figure}[t]
\centering
\includegraphics[width=0.5\textwidth]{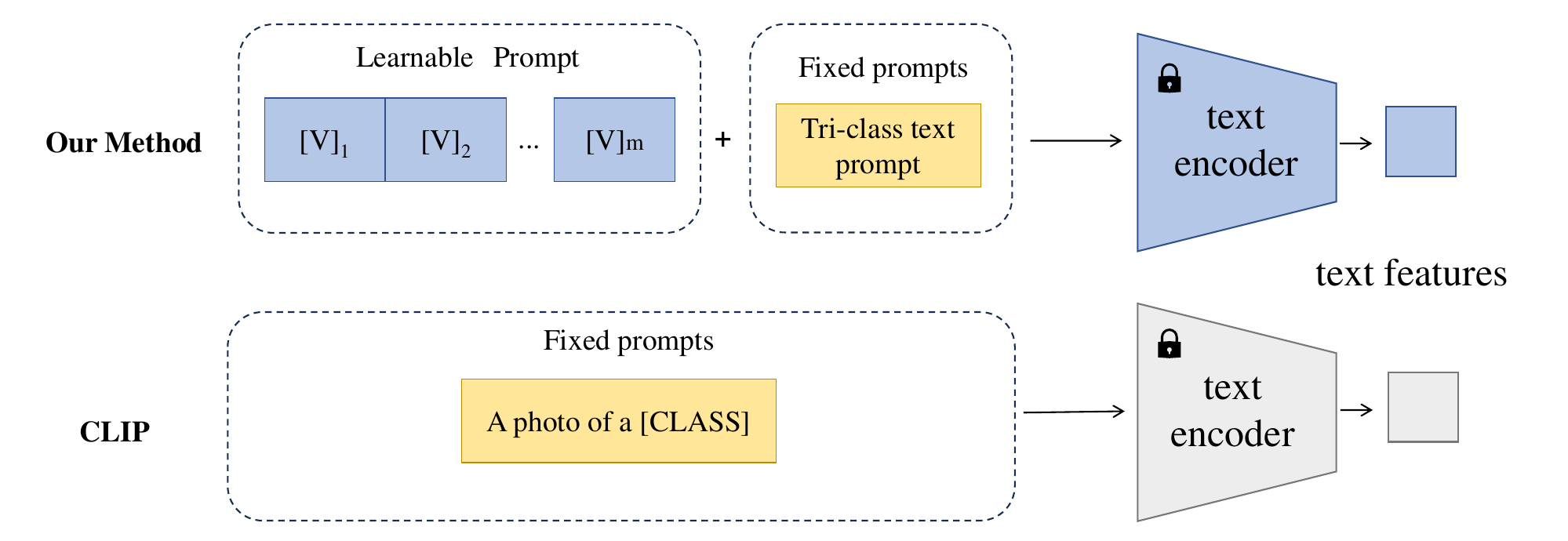}
\caption{The different between our method and CLIP in the Text Encoder.}
\label{fig-main2}
\end{figure}
The image encoder from CLIP, denoted as $A(\cdot)$, transforms an input image $x \in R^{3\times H\times W}$, where $H$ and $W$ represent the height and width dimensions, respectively, into a $d$-dimensional image feature vector $A_{x} \in R^{N\times d}$. Here, $N$ signifies the number of patches into which the image is divided, and $d$ is the dimensionality of the feature space.
To enhance the image features with domain-specific remote sensing ship priors, we incorporate the image $x$ into a specialized remote-sensing encoder, denoted as $B(\cdot)$. This encoder, analogous to CLIP's image encoder $A(\cdot)$, processes the image $x$ into a feature vector $B_{x} \in R^{N\times d}$, maintaining the same patch division as before.
These remote-sensing ship priors, represented by $B_{x}$, are refined by Visual-Net $Net(\cdot)$, a lightweight neural network. The remote-sensing bias, derived from this network, is then combined with the features from CLIP's image encoder to form the final image feature representation $I_{x}$:
\begin{equation}
I_{x} = A_{x} + Net(B_{x}) \label{eq1}.
\end{equation}
As a result, the final image features $I_{x}$ encapsulate not only the rich knowledge encoded by CLIP but also the nuanced domain insights from remote sensing ship priors. These priors are effectively integrated through bias terms learned by the small, yet effective, trainable networks, enhancing the model's ability to generalize and adapt to novel ship classes in the remote sensing context.

In our CLIP implementation, we use the officially open-sourced ViT-B/16 and ResNet-50 models. For the RS-encoder, we use a pre-trained remote sensing-specific ViT as the encoder, and freeze this part to prevent gradient backpropagation. Specifically, we use the ViT-B/16 pre-trained on the MillionAID \cite{millionaid} dataset as the RS-encoder \cite{rvsa,selectivemae}.

This choice has several advantages: Firstly, compared to the ViT-B/16 pre-trained on ImageNet, the ViT-B/16 pre-trained on MillionAID has a significant remote sensing bias, allowing it to better fit remote sensing domain knowledge of ships. Secondly, the same ViT-B/16 architecture makes it easier to align the knowledge between the general CLIP end and the remote sensing domain. In summary, we chose the ViT-B/16 pre-trained on MillionAID as the RS-encoder. Although we have achieved good results, we believe that further improvements in remote sensing pre-trained ViT and better alignment with the CLIP encoder should yield better performance gains. We will continue to explore this in the future.

\subsection{Multi-granularity prompt}
CLIP \cite{clip} typically employs hand-crafted prompts, such as 'a photo of a [ ]', to generate textual embeddings. Our approach augments these hand-crafted prompts with learnable text prompts, designed to capture the rich details present in ship images. Figure \ref{fig-main2} shows the differences between our method and CLIP.

To refine the prompts, we introduce a hierarchical, multi-granularity prompt design. The class token embeddings $c_i$ for the $i$-th class are structured across three levels of granularity and are modified accordingly:
\begin{equation}
c_i = \{c_i^{p}, c_i^{s}, c_i^{f}\},
\end{equation}
where $c_i^{p}, c_i^{s}, c_i^{f}$ denote the primary type, secondary type, and final type of $c_i$, respectively.
The hand-crafted prompts are tailored to align with these class token embeddings $c_i$ and are modified as
$t_i$ = \{\text{a photo of a ship, the primary type is } $c_i^{p}$, \text{ secondary type is } $c_i^{s}$, \text{ final type is } $c_i^{f}$\}.
Building upon this hierarchical prompt, we introduce a set of $M$ context vectors $V = \{v_1, v_2, ..., v_M\}$ as learnable text prompts. These learnable vectors $V$ are combined with a remote bias to create new context tokens. The remote bias is generated by a remote-sensing encoder that transforms the image $x$ into a $d$-dimensional feature vector $B_x \in R^{N \times d}$.
A lightweight neural network, named Remote-Net, parameterized by $\theta$, processes the feature $B_x$ to generate a remote-sensing text bias vector. This bias vector is then merged with each learnable text context vector:
\begin{equation}
     v_m(x) = v_m + \delta,
\end{equation}
where $\delta = r_\theta(x)$ and $m \in \{1, 2, ..., M\}$.
In this way, the new learnable vectors $V$ not only incorporate the knowledge encoded by CLIP but also integrate remote sensing ship priors.

The class token embeddings $c_i$ and their corresponding hierarchical classes $t_i$ are effectively integrated with the adaptable context vector $V$:
\begin{equation}
p_i = \{v_1(x), v_2(x), ..., v_M(x), t_i\}.
\end{equation}

The adaptable text prompt $p_i$ is then fed into the text encoder $T(\cdot)$, resulting in the textual class embedding $T(p_i)$.
The image encoder processes the image $x$ to extract the image feature $I_x$ and computes it with the text feature $T(p)$ derived from the new learnable prompt. The goal is to classify the image into one of the $C$ unique classes, represented by $z \in \{1, ..., C\}$, where $z$ is the predicted class label.
The predicted probability of the $i$-th class is expressed as:
\begin{equation}
P(z = i | x) = \frac{\exp(\text{cos}(I(x), T(p_i))/\tau)}{\sum_{j=1}^{C}\exp(\text{cos}(I(x), T(p_j))/\tau)},
\end{equation}
where $\text{cos}(\cdot, \cdot)$ and $\tau$ denote the cosine similarity and the temperature parameter of the softmax function, respectively. The die

During training, we simultaneously update the context vectors, Remote-Net parameters, and Visual-Net parameters. Both Remote-Net and Visual-Net are implemented as two-layer bottleneck networks (\textit{i.e.}, Linear-ReLU-Linear), reducing the input dimension by 16 in the hidden layer. The inputs to these networks are derived from the outputs of the remote-sensing encoder.
We leave the exploration of more advanced network architectures as future work.

\section{Introduction of FGSCM-52}

\begin{figure}[t]
\centering
\includegraphics[width=0.5\textwidth]{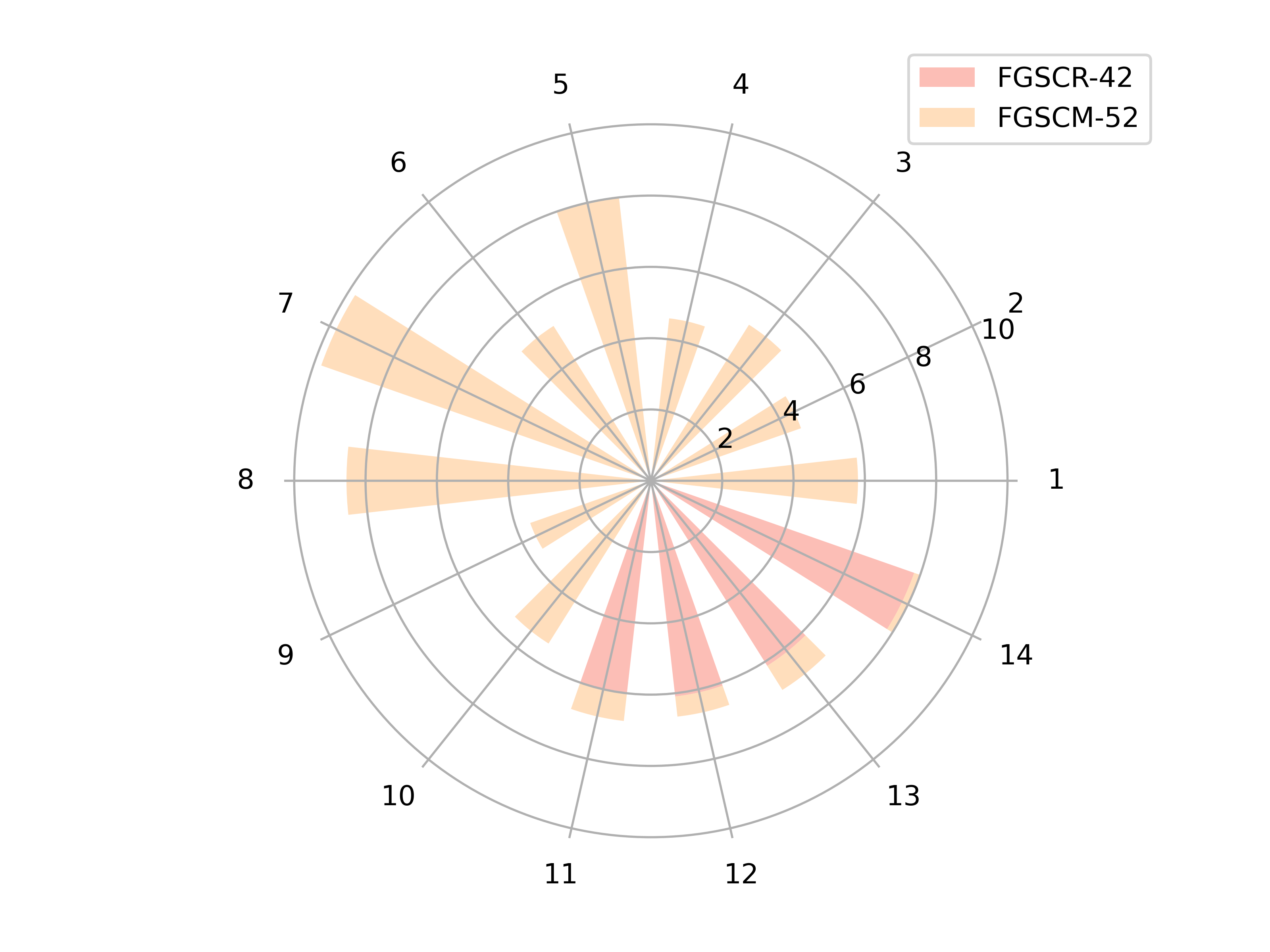}
\caption{Overview of the difference between our FGSCM-52 dataset and the original FGSCR-42 dataset. Sample quantities are presented on a logarithmic scale.}
\label{fig4}
\end{figure}
The development of the RS-FGSC field necessitates large-scale realistic fine-grained ship datasets. The FGSC-23 \cite{fgsc23} and FGSCR-42 \cite{fgsc42} datasets have been pivotal for practical applications, yet they exhibit limitations when assessing advanced classification methods.
Specifically, FGSC-23 categorizes data into relatively broad classes, such as lumping all auxiliary ships into one category, which does not align with the granularity required in real-world scenarios. Additionally, the dataset's limited number of categories fails to satisfy the diverse needs of practical applications.
FGSCR-42, while providing finer classifications within some broader categories, still lacks comprehensive subcategory coverage. Moreover, the dataset suffers from an imbalance in sample distribution, with six categories containing fewer than ten samples each, which can hinder effective model training.

To overcome these limitations, we introduce the FGSCM-52 dataset, an enhanced version of FGSCR-42, designed to offer a more extensive collection of fine-grained remote sensing ship images. FGSCM-52 includes an additional ten categories and increases the number of samples across existing categories, thereby making it serve as a better testbed for fine-grained ship classification.
The dataset features image sizes ranging from 50$\times$50 to 1,600$\times$1,700 pixels. 
Furthermore, FGSCM-52 is equipped with well-defined hierarchical labels, which facilitate a more nuanced understanding of ship classifications. For a comprehensive overview of the FGSCM-52 dataset, please refer to Fig. \ref{fig4}.

\section{Experiments}
Following the CLIP and CoOp, we evaluate the model's base-to-new generalization ability across various datasets and settings.
To validate the effectiveness of base-to-new generalization in the RS-FGSC field, we compare our method with other methods under different settings.
Additionally, we conducted comprehensive ablation studies to investigate the impact of each component in our method. 
To further analyze the ability of our method, we also employ EuroSAT \cite{eurosat} to conduct further comparative experiments following CoOp.

\subsection{Datasets}
Two publicly available datasets are employed for fine-grained ship classification, including FGSC-23 \cite{fgsc23} and FGSCR-42 \cite{fgsc42}.
Furthermore, we also use the proposed FGSCM-52 dataset, which is more challenging in the base-to-new setting.
Publicly available datasets such as the EuroSAT \cite{eurosat} are also employed for further experiments.
In the RS-FGSC field, we split the fine-grained ship classification datasets for the base-to-new generalization tasks, following the practice in CLIP and CoOp methods.

\subsubsection{FGSC-23}
The FGSC-23 dataset, derived from public Google Earth data and the GF-2 satellite, comprises 4,052 images of ship targets.
With image resolutions varying from 0.4-2m, each slice aligns the target with the image center.
The dataset includes 23 different classes, one of which represents objects resembling ships but not ships, while the other 22 are real ship targets.
Nevertheless, FGSC-23 lacks sufficient details and abundant categories, impairing its effectiveness in fine-grained ship classification tasks.
In our experiments, we annotate multi-level hierarchical labels for FGSCR-42.

\subsubsection{FGSCR-42}
The remote sensing images in the FGSCR-42 dataset, mainly come from Google Earth and the previous datasets, like FGSC-23.
The entire FGSCR-42 dataset includes 9,320 images across 42 categories.
Image sizes in FGSCR-42 vary from approximately 50 $\times$ 50 pixels to around 1,500 $\times$ 1,500 pixels.
In FGSCR-42, six categories have fewer than ten samples each.
Considering the special setting in base-to-new generalization, we have excluded those categories with fewer than ten samples for new tasks.
In our experiments, we annotate multi-level hierarchical labels for FGSCR-42.

\subsubsection{EuroSAT}
EuroSAT is a land use and land cover classification dataset derived from the Sentinel-2 satellite. 
It stands out from previous datasets due to its extensive spectral coverage, including thirteen spectral bands across visible, near-infrared, and short-wave infrared spectra.
The dataset contains 27,000 georeferenced and annotated images, divided into ten categories: Industrial Buildings, Residential Buildings, Annual Crop, Permanent Crop, River, Sea or Lake, Herbaceous Vegetation, Highway, Pasture, and Forest.
Each category features 2,000 to 3,000 remote sensing images, with every image measuring 64 $\times$ 64 pixels.
Although it is a typical remote-sensing dataset, it contains no ship images.
Therefore, it is used to evaluate the effectiveness of the proposed method for related tasks.

\subsection{Implementation details}
\subsubsection{Baselines}
Two types of CLIP-based methods are treated as baselines for comparison in the base-to-new generalization setting:
\begin{itemize}
\item CoOp (IJCV'22 \cite{coop}) enhances the CLIP model for the base-to-new generalization task by replacing hand-crafted text prompts with learnable text prompts. 
\item CoCoOp (CVPR'22 \cite{cocoop}) adds learnable parameters to the image aspect, creating image-conditional prompts that merge each image's context with learnable prompts. This collaborative fine-tuning of text and images enables CoCoOp to excel in base-to-new generalization tasks.
\end{itemize}

\subsubsection{Training details}
Comparison methods (CoOp and CoCoOp) fix context length at 16 without initializing context vectors, positioning the class token at the end. 
Following CoOp, we also use common vision backbones, ResNet-50 \cite{resnet} and ViT-B/16 \cite{vit}, and a K-shot setting in base-to-new generalization, with K values of $\{$1,2,4,6$\}$.
Especially, to make a fair comparison with other CoOp-based methods, we use the K values of $\{$4,8$\}$, which has been well reported in previous works.
The hyperparameters are set as below, \textit{e.g.}, training for 100 epochs and using the SGD optimizer, a 1e-3 learning rate, a MultiStepLR learning rate scheduler, and a 1e-4 weight decay for L2 regularization.
Except for accuracy on base and new classes, we also use the harmonic mean (denoted as H) of the accuracy on base and novel classes to represent the overall ability of the different models, which is widely used in the CoOp-based methods.

\begin{table*}[!t]
\caption{Comparison with different methods in different K-shot settings on the FGSC-23 and FGSCR-42 datasets with different backbones (ViT-B/16 and ResNet-50). H: Harmonic mean.}
\centering
\renewcommand\arraystretch{1.2}
\resizebox{1.0\linewidth}{!}{
\begin{tabular}{@{}>{\raggedright\arraybackslash}p{1.5cm}>{\raggedright\arraybackslash}p{1.5cm}>{\raggedright\arraybackslash}p{1.5cm}
>{\raggedright\arraybackslash}p{1.5cm}cccccccc@{}}
\toprule
\multicolumn{3}{c}{Method} & \multicolumn{3}{c}{CoOp} & \multicolumn{3}{c}{CoCoOp} & \multicolumn{3}{c}{Ours} \\
\cmidrule(r){1-3} \cmidrule(lr){4-6} \cmidrule(l){7-9} \cmidrule(l){10-12}
Dataset & Backbone & Shots & Base & New & H & Base & New & H & Base & New & H \\
\midrule
\multirow{8}{*}{FGSC-23} & \multirow{4}{*}{ViT-B/16} & K=1 & 22.67 $\pm$ \tiny{0.88} & 18.30 $\pm$ \tiny{7.4} & 20.25 & 16.43 $\pm$\tiny{0.84} & 13.67 $\pm$ \tiny{1.75} & 14.92 & 23.30 $\pm$ \tiny{3.43} & 19.70$\pm$\tiny{0.91} & 21.35 \\
                        & & K=2 & 25.47 $\pm$ \tiny{0.95} & 15.53 $\pm$ \tiny{0.33} & 19.30 & 15.67 $\pm$ \tiny{0.82} & 15.30 $\pm$ \tiny{1.71}& 15.48 & 31.37 $\pm$ \tiny{1.10} & 17.40 $\pm$ \tiny{1.84} & 22.38 \\
                        & & K=4 & 32.40 $\pm$ \tiny{0.78} & 18.30 $\pm$ \tiny{4.01}& 23.39 & 16.37 $\pm$ \tiny{0.41} & 13.90 $\pm$ \tiny{2.06}& 15.03 & 37.93 $\pm$ \tiny{1.64} & 23.50$\pm$\tiny{0.89} & 29.02 \\
                        & & K=6 & 30.50 $\pm$ \tiny{0.99} & 16.90 $\pm$ \tiny{3.65}& 21.75 & 19.27 $\pm$ \tiny{2.21}& 13.67 $\pm$ \tiny{1.75}& 15.99 & 43.63 $\pm$ \tiny{3.84} & 17.85$\pm$\tiny{0.49} & 25.33 \\
\cline{2-12} 
\addlinespace[2pt] 
                        & \multirow{4}{*}{ResNet-50} & K=1 & 15.70 $\pm$ \tiny{2.33}& 15.65 $\pm$ \tiny{1.75}& 15.67 & 14.87 $\pm$ \tiny{1.03}& 16.43 $\pm$ \tiny{2.65}& 15.61 & 20.00 $\pm$ \tiny{1.16} & 17.83 $\pm$ \tiny{6.12} & 11.25 \\
                        & & K=2 & 19.60 $\pm$ \tiny{0.28}& 19.47 $\pm$ \tiny{6.00}& 19.53 & 12.97 $\pm$ \tiny{1.92}& 16.90 $\pm$ \tiny{4.01}& 14.68 & 24.60 $\pm$ \tiny{2.87} & 19.80 $\pm$ \tiny{7.01} & 21.94 \\
                        & & K=4 & 25.17 $\pm$ \tiny{1.14}& 15.30 $\pm$ \tiny{2.62}& 19.03 & 19.50 $\pm$ \tiny{2.42}& 13.20 $\pm$ \tiny{2.97}& 15.74 & 29.03 $\pm$ \tiny{0.09} & 20.37 $\pm$ \tiny{2.90} & 23.94 \\
                        & & K=6 & 29.37 $\pm$ \tiny{1.56}& 16.70 $\pm$ \tiny{1.51}& 21.29 & 20.93 $\pm$ \tiny{2.24}& 11.33 $\pm$ \tiny{1.75}& 14.70 & 33.70 $\pm$ \tiny{2.05} & 25.00 $\pm$ \tiny{6.28} & 28.71 \\
\midrule
\multirow{8}{*}{FGSCR42} & \multirow{4}{*}{ViT-B/16} & K=1 & 36.23 $\pm$ \tiny{4.73} & 9.40 $\pm$ \tiny{1.84}& 14.93 & 21.90$\pm$ \tiny{2.20} & 16.93 $\pm$ \tiny{2.28} & 19.10 & 45.30 $\pm$ \tiny{4.00} & 17.43 $\pm$ \tiny{0.93} & 25.17 \\
                         & & K=2 & 41.17 $\pm$ \tiny{5.85} & 17.47 $\pm$ \tiny{4.21}& 24.53 & 41.37 $\pm$ \tiny{2.02} & 14.77 $\pm$ \tiny{3.41}& 21.77 & 57.53 $\pm$ \tiny{1.86} & 15.20$\pm$\tiny{3.62} & 22.47 \\
                         & & K=4 & 35.57 $\pm$ \tiny{3.22} & 16.07 $\pm$ \tiny{4.32}& 22.14 & 42.11 $\pm$ \tiny{2.01} & 16.50 $\pm$ \tiny{2.92} & 23.71 & 67.67 $\pm$ \tiny{4.24} & 19.07 $\pm$ \tiny{2.22} & 29.75 \\
                         & & K=6 & 56.4 $\pm$ \tiny{1.07} & 13.23 $\pm$ \tiny{2.82}& 21.43 & 60.93 $\pm$ \tiny{0.52} & 13.23 $\pm$ \tiny{2.09}& 21.74 & 73.57 $\pm$ \tiny{3.03} & 20.90 $\pm$ \tiny{0.82} & 32.55 \\
\cline{2-12} 
\addlinespace[2pt] 
                         & \multirow{4}{*}{ResNet-50} & K=1 & 25.17 $\pm$ \tiny{1.30}& 11.50 $\pm$ \tiny{1.90}& 15.79 & 15.43 $\pm$ \tiny{2.16}& 11.10 $\pm$ \tiny{4.67}& 12.91 & 42.43 $\pm$ \tiny{3.16}& 18.50 $\pm$ \tiny{2.09}& 25.77 \\
                         & & K=2 & 25.90 $\pm$ \tiny{8.66}& 18.13 $\pm$ \tiny{2.47}& 21.33 & 22.87 $\pm$ \tiny{2.59}& 15.73 $\pm$ \tiny{4.21}& 18.64 & 54.57 $\pm$ \tiny{2.64}& 18.90 $\pm$ \tiny{1.59}& 28.08 \\
                         & & K=4 & 36.20 $\pm$ \tiny{5.80}& 17.60 $\pm$ \tiny{3.75}& 23.68 & 34.43 $\pm$ \tiny{3.73}& 10.50 $\pm$ \tiny{3.51}& 16.09 & 64.65 $\pm$ \tiny{2.95}& 19.90 $\pm$ \tiny{0.40}& 30.43 \\
                         & & K=6 & 49.57 $\pm$ \tiny{1.96}& 22.47 $\pm$ \tiny{3.43}& 30.92 & 41.23 $\pm$ \tiny{4.26}& 13.13 $\pm$ \tiny{0.96}& 19.92 & 75.73 $\pm$ \tiny{1.55}& 21.10 $\pm$ \tiny{3.29}& 27.90 \\
\bottomrule
\end{tabular}
}
\label{table1}
\end{table*}

\begin{table*}[!t]
\caption{Comparison with different methods in different K-shot settings on the proposed FGSCR-52 dataset with different backbones (ViT-B/16 and ResNet-50). H: Harmonic mean.}
\centering
\renewcommand\arraystretch{1.1}
\resizebox{1.0\linewidth}{!}{
\begin{tabular}{@{}lcccccccccc@{}}
\toprule
\multirow{2}{*}{Backbones} & \multirow{2}{*}{Shots} & \multicolumn{3}{c}{CoOp} & \multicolumn{3}{c}{CoCoOp} & \multicolumn{3}{c}{Ours} \\
\cmidrule(r){3-5} \cmidrule(lr){6-8} \cmidrule(l){9-11}
&   & Base & New & H & Base & New & H & Base & New & H \\ \midrule
\multirow{4}{*}{ViT-B/16} 
& K=1 & 5.53$\pm$ \tiny{1.96} & 1.90 $\pm$ \tiny{0.54}& 2.83 & 15.53$\pm$ \tiny{1.96} & 10.00$\pm$ \tiny{2.97} & 12.17 & 30.13 $\pm$ \tiny{3.62}& 12.60 $\pm$ \tiny{3.26}& 17.77 \\
& K=2 & 6.50 $\pm$ \tiny{2.24}& 17.47 $\pm$ \tiny{3.84}& 9.47 & 26.5 $\pm$ \tiny{2.74}& 7.10$\pm$ \tiny{4.20} & 11.20 & 46.37 $\pm$ \tiny{2.21}& 11.30 $\pm$ \tiny{2.60}& 18.17 \\
& K=4 & 12.33$\pm$ \tiny{2.37} & 9.10 $\pm$ \tiny{3.26}& 10.47 & 42.33 $\pm$ \tiny{2.37}& 5.20$\pm$ \tiny{0.54} & 9.26 & 57.77 $\pm$ \tiny{4.25}& 14.20 $\pm$ \tiny{1.27}& 22.80 \\
& K=6 & 21.07 $\pm$ \tiny{0.78}& 6.95$\pm$ \tiny{2.25} & 10.45 & 51.07 $\pm$ \tiny{0.78}& 13.23$\pm$ \tiny{10.18} & 21.02 & 68.30$\pm$ \tiny{0.12} & 15.20 $\pm$ \tiny{1.50}& 24.87 \\
\cmidrule{1-11} 
\multirow{4}{*}{ResNet-50} 
& K=1 & 8.47 $\pm$ \tiny{1.96}& 9.67$\pm$ \tiny{0.24} & 9.03 & 20.60 $\pm$ \tiny{2.55}& 4.27$\pm$ \tiny{3.94} & 7.07 & 32.83$\pm$ \tiny{4.52} & 13.20 $\pm$ \tiny{4.87}& 18.83 \\
& K=2 & 13.40$\pm$ \tiny{1.69} & 12.10$\pm$ \tiny{1.31} & 12.72 & 20.40$\pm$ \tiny{11.06} & 5.50$\pm$ \tiny{1.60} & 8.66 & 46.80$\pm$ \tiny{2.70} & 14.57 $\pm$ \tiny{3.27}& 22.22 \\
& K=4 & 26.63 $\pm$ \tiny{5.41}& 3.97$\pm$ \tiny{2.20} & 6.91 & 42.87$\pm$ \tiny{3.82} & 11.77$\pm$ \tiny{7.01} & 18.47 & 58.33 $\pm$ \tiny{3.12}& 16.21$\pm$ \tiny{2.13} & 25.37 \\
& K=6 & 37.83 $\pm$ \tiny{3.63}& 10.27 $\pm$ \tiny{7.95}& 16.15 & 50.03$\pm$ \tiny{2.93} & 6.03$\pm$ \tiny{2.00} & 10.76 & 64.85$\pm$ \tiny{2.55} & 11.71 $\pm$ \tiny{1.03}& 19.84 \\ 
\bottomrule
\end{tabular}
}
\label{table2}
\end{table*}

\subsection{Base-to-New Generalization}
Similar to other CLIP-based methods, we divide the ship dataset into base classes for training and new classes for testing. 
We validated the model's performance on both base classes and new classes.
In particular, we first fine-tuned all CLIP-based methods on base classes and then tested them on the new classes.
The main results on base and new classes are summarized in Table \ref{table1}. 

As can be seen, compared to existing CLIP-based works, our method outperforms them on two open remote sensing fine-grained ship datasets.
Remarkably, compared with CoCoOp, our method exhibits significant advancement in base and new classes, notably reaching a harmonic mean of 25.33 and 33.76 in the 6-shot setting with ViT-B/16 as the backbone, which considerably exceeds CoCoOp. In particular, our method achieves the accuracy of 43.63$\%$ and 73.57$\%$ in the 6-shot setting with ViT-B/16 as the backbone on the base classes, which considerably exceeds CoCoOp's accuracy. Moreover, our method also improves accuracy in new classes when compared to CoOp and CoCoOp. For instance, in the 6-shot setting on the FGSC-23 and FGSCR-42 dataset with ViT-B/16 backbone, our method exceeds CoCoOp by 5.58$\%$ and 8.68$\%$, respectively. 

Notably, our approach becomes more effective as the number of categories increases, with the performance improving in correlation with the increasing number of classes across the FGSC-23 and FGSCR-42 datasets.
For instance, in the 6-shot setting with ViT-B/16 as the backbone, our method outperforms CoCoOp by the harmonic mean of 9.34 and 10.81. 
The improved performance is less obvious in the FGSC-23 dataset with fewer categories compared to the FGSCR-42 datasets.

Additionally, CLIP-based fine-tuning methods perform differently when using different backbones.
For instance, CLIP-based models using ResNet-50 as the backbone generally fall short compared to those using ViT-B/16.
Specifically, in the 1-shot setting on the two datasets, using ViT-B/16 instead of ResNet-50, our method achieves a gain of 10.1 and 0.71 in terms of the harmonic mean, respectively. 

Generally, the performance increases when more data are used.
However, on the FGSC-23 dataset, there's a notable decrease in accuracy in the 6-shots setting.
This drop is likely due to the limited number of categories in the FGSC-23 dataset, as it's not observed in datasets with more categories, like FGSCR-42. In summary, our proposed method has been proven to enhance the model's ability in base and new classes, thereby establishing itself as a strong baseline for the base-to-new generalization tasks in the RS-FGSC field.

\subsection{Comparisons on the new FGSCM-52 dataset}
To further analyze the efficacy of our method on the base-to-new generalization task, we compare our method with other methods on the new FGSCM-52 dataset.
The new FGSCM-52 dataset, which contains more categories in the RS-FGSC field, presents more challenges than other open-ship datasets, as evidenced by the inferior performance of the CLIP-based methods.
We conducted extensive experiments following the previous base-to-new generalization task settings, with results shown in Table \ref{table2}.

\begin{figure*}[t!]
    \centering 	
    \subfloat[]{\includegraphics[width=0.49\textwidth]{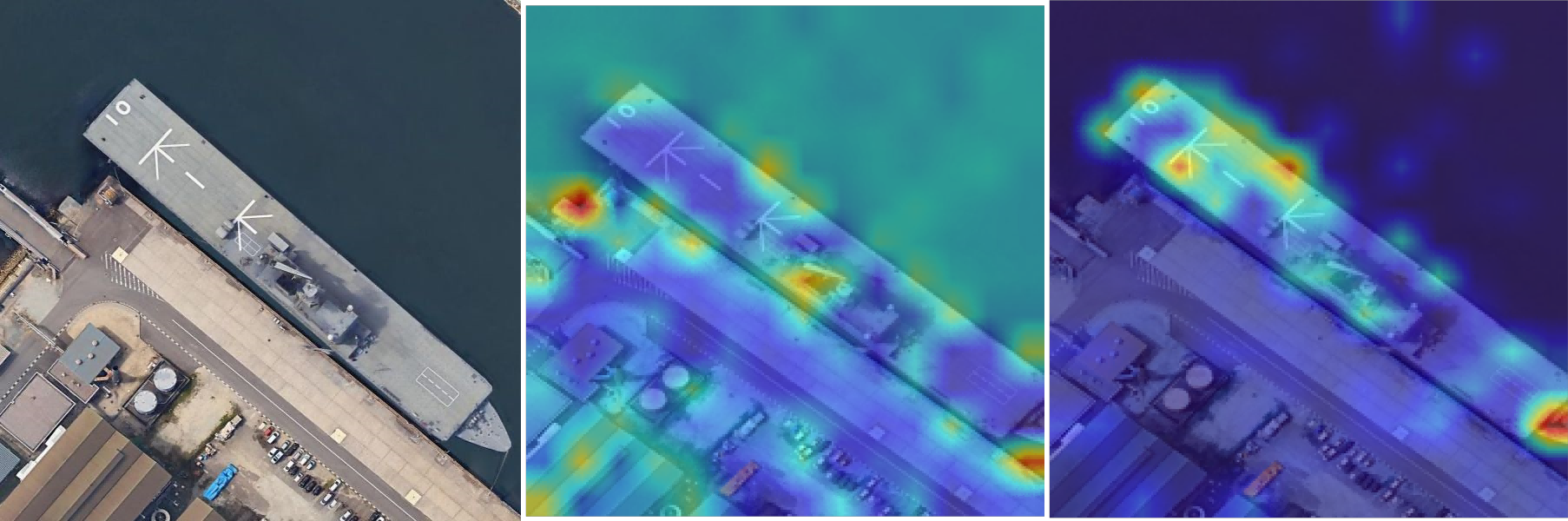}}
    \hfill 	
    \subfloat[]{\includegraphics[width=0.49\textwidth]{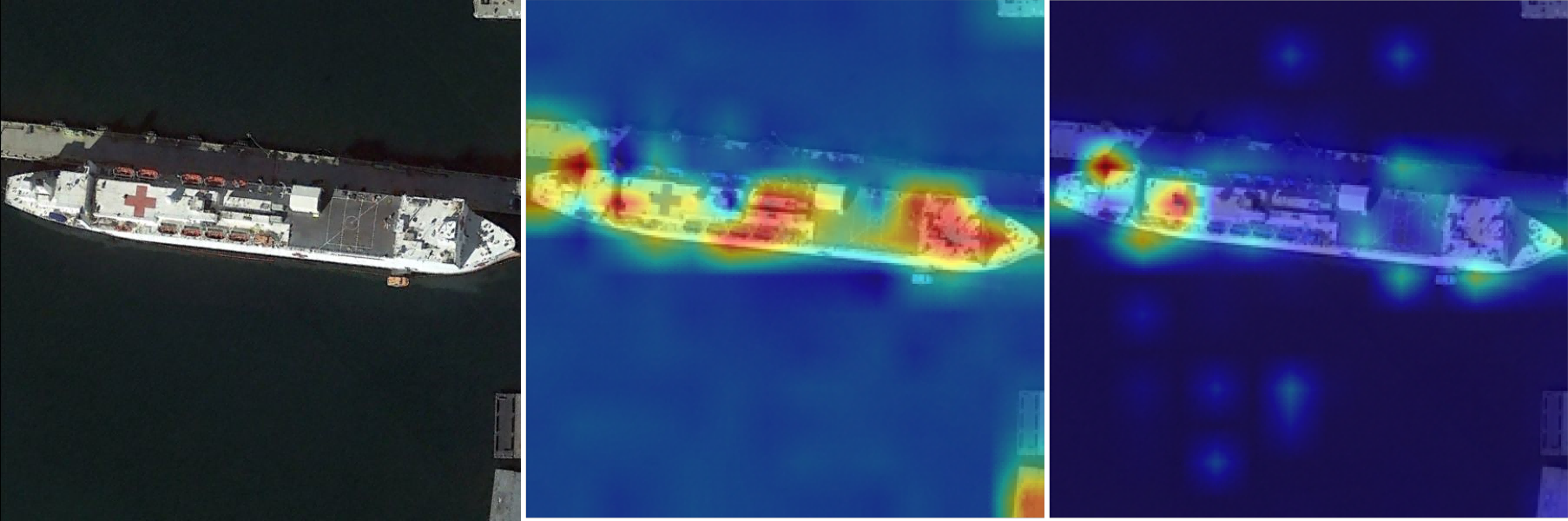}}
    \newline
    \subfloat[]{\includegraphics[width=0.49\textwidth]{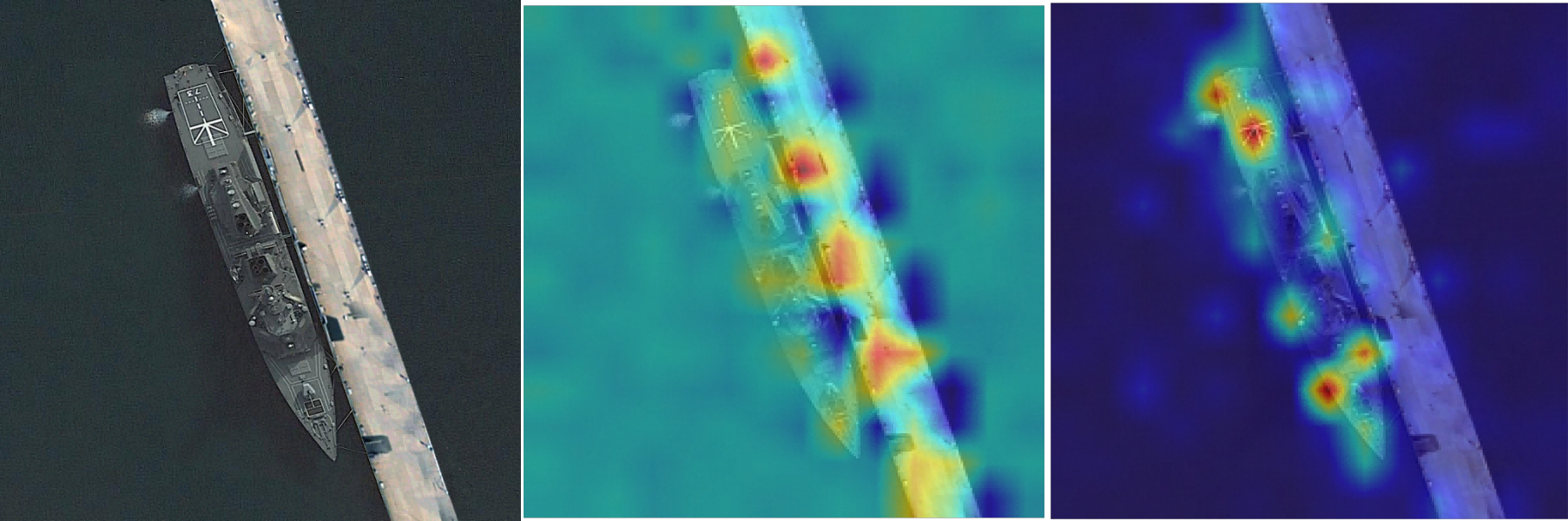}}
    \hfill 	
    \subfloat[]{\includegraphics[width=0.49\textwidth]{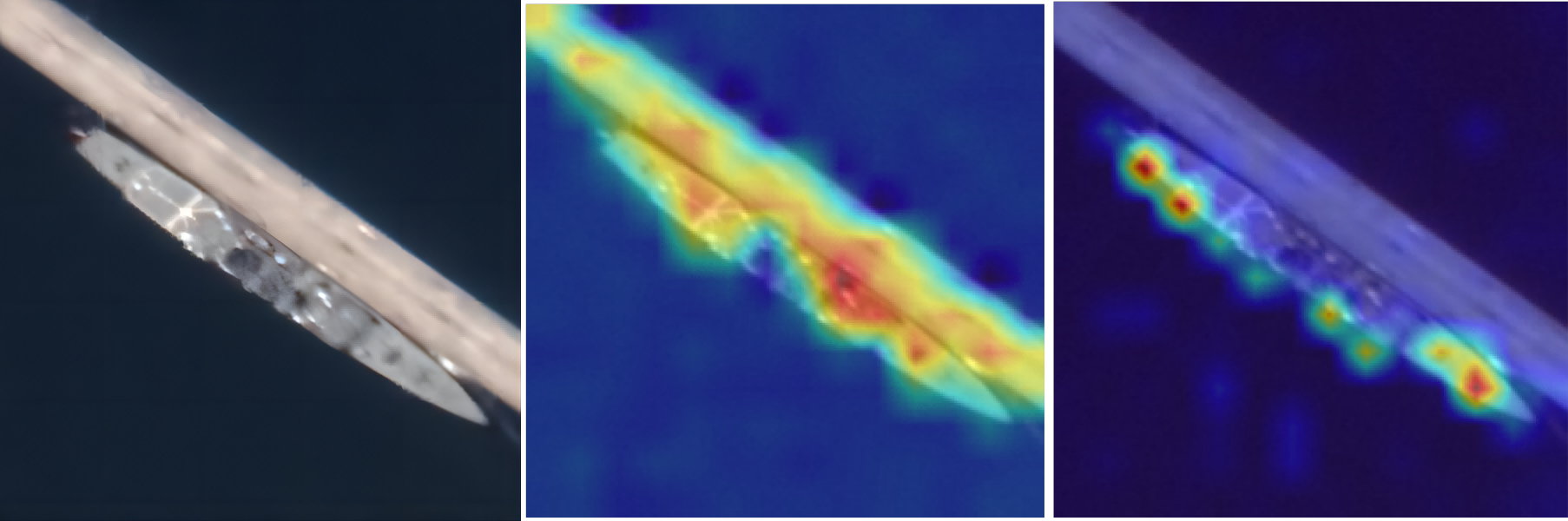}}
    \caption{Visualization of the attention map of different methods. In each group, the ship image, the attention map of the CLIP method, and the attention map of the proposed method are displayed from left to right.}
    \label{fig5}
\end{figure*}

Analysis of Table \ref{table2} reveals that our method, which uses the multi-granularity prompts and remote sensing ship priors, achieves competitive results compared with the CLIP-based works.
It significantly outperforms other VLM fine-tuning techniques like CoOp and CoCoOp in the RS-FGSC field, achieving an impressive 14.42 and 3.85 improvement of the harmonic mean over CoOp and CoCoOp in the 6-shot setting with ViT-B/16 as the backbone.
It is noteworthy that the overall performance on the FGSCM-52 dataset is not as good as that on the FGSCR-42 dataset, suggesting the difficulty of the new FGSCM-52 dataset. The FGSCM-52 dataset offers more categories and poses a new challenge for the base-to-new generalization task in the RS-FGSC field.
Nevertheless, our method shows a solid baseline for future research and marks a pioneering step in applying VLMs in the RS-FGSC field.

\begin{table*}[!t]
\caption{
Comparison with different methods in different K-shot settings on the EuroSAT dataset with different backbones (ViT-B/16 and ResNet-50). H: Harmonic mean.}
\centering
\renewcommand\arraystretch{0.9}
\begin{tabular}{@{}>{\raggedright\arraybackslash}p{1.5cm}>{\raggedright\arraybackslash}p{1.2cm}
>{\raggedright\arraybackslash}p{0.8cm}cccccccc@{}}
\toprule
    \multirow{2}{*}{Backbones} & \multirow{2}{*}{Shots} & \multicolumn{3}{c}{CoOp} & \multicolumn{3}{c}{CoCoOp} & \multicolumn{3}{c}{Ours} \\
    \cmidrule(r){3-5} \cmidrule(r){6-8} \cmidrule(r){9-11}
     &  & Base & New & H & Base & New & H & Base & New & H \\
    \midrule
    \multirow{2}{*}{ViT-B/16} & K=4 & 82.56 & 53.04 & 64.59 & 79.27 & 65.44 & 71.69 & 85.27 & 66.50 & 74.72 \\
     & K=8 & 83.27 & 50.59 & 62.94 & 78.68 & 56.03 & 65.45 & 86.09 & 62.23 & 72.24 \\
    \midrule 
    \multirow{2}{*}{ResNet-50} & K=4 & 86.39 & 46.91 & 60.80 & 75.60 & 37.68 & 50.29 & 87.52 & 49.12 & 62.92 \\
     & K=8 & 85.88 & 42.46 & 56.83 & 80.43 & 48.75 & 60.71 & 88.34 & 54.67 & 67.54 \\
    \bottomrule
  \end{tabular}
\label{table3}
\end{table*}

\subsection{Comparisons on other datasets}
Our approach, specifically designed for the RS-FGSC domain, has notably improved performance, as evidenced in Section V.C. To comprehensively compare our method's effectiveness with other CLIP-based methods, we also conducted experiments across datasets from different fields. We selected these datasets with two criteria in mind: these datasets must be extensively used in CLIP-based research and bear similarities to our task for a relevant comparison. Initially, we opted for the EuroSAT dataset, rich in remote sensing data and closely related to our objectives despite not being specific to remote sensing of ships.

The results are presented in Table \ref{table3}, demonstrating our method's superior performance across datasets. Specifically, in a 4-shot and 8-shot setting with the ViT-B/16 backbone, our method outperformed CoCoOp on the EuroSAT dataset, achieving a gain of 3.03 and 6.79 in terms of harmonic mean, respectively. Moreover, it showed remarkable feature extraction ability in base classes, surpassing CoCoOp by 6.0 and 7.41 in terms of harmonic mean, respectively. Meanwhile, it achieved performance boosts in new classes.

In summary, the additional experiments across various datasets which are quite different from the datasets in RS-FGSC domains validate our method's superiority and its broad applicability.
\begin{table*}[t!]
\caption{Ablation study results of the proposed method on the FGSCR-42 dataset with the ViT-B/16 backbone.}
\centering
\renewcommand\arraystretch{1.1}
\resizebox{0.8\linewidth}{!}{
\begin{tabular}{c|ccc|ccc}
\hline
\multirow{2}{*}{Shots} & \multicolumn{3}{c|}{Methods} & \multirow{2}{*}{Base} & \multirow{2}{*}{New} & \multirow{2}{*}{H} \\ \cline{2-4}
 & baseline & \begin{tabular}[c]{@{}c@{}}multi-class text prompt\end{tabular} & \begin{tabular}[c]{@{}c@{}}remote-sensing bias\end{tabular} &  & & \\ \hline
\multirow{4}{*}{1} & $\surd $ &  &  & 21.90$\pm$\tiny{2.20} & 16.93$\pm$\tiny{2.28} & 19.10 \\
 & $\surd $ & $\surd $ &  & 31.50$\pm$\tiny{3.57} & 17.17$\pm$\tiny{1.73} & 21.13 \\
 & $\surd $ &  & $\surd $ & 37.05$\pm$\tiny{3.48} & 17.21$\pm$\tiny{1.13} &  23.41\\
 & $\surd $ & $\surd $ & $\surd $ & 45.30$\pm$\tiny{4.00} & 17.43$\pm$\tiny{0.93} & 25.17 \\ \hline
\multirow{4}{*}{2} & $\surd $ &  &  & 41.37$\pm$\tiny{2.02} & 14.77$\pm$\tiny{3.41} & 21.77 \\
 & $\surd $ & $\surd $ &  & 47.38$\pm$\tiny{2.86} & 12.96$\pm$\tiny{3.48} & 21.58 \\
 & $\surd $ &  & $\surd $ & 49.32$\pm$\tiny{1.74} & 15.58$\pm$\tiny{1.67} & 21.89 \\
 & $\surd $ & $\surd $ & $\surd $ & 57.53$\pm$\tiny{1.86} & 15.20$\pm$\tiny{3.62} & 22.47 \\ \hline
\multirow{4}{*}{4} & $\surd $ &  &  & 42.11$\pm$\tiny{2.01} & 16.50$\pm$\tiny{2.92} & 23.71 \\
 & $\surd $ & $\surd $ &  & 54.63$\pm$\tiny{1.07} & 17.10$\pm$\tiny{1.72} &  26.17\\
 & $\surd $ &  & $\surd $ & 55.60$\pm$\tiny{0.95} & 18.88$\pm$\tiny{1.67} &  27.18\\
 & $\surd $ & $\surd $ & $\surd $ & 67.67$\pm$\tiny{4.24} & 19.07$\pm$\tiny{2.23} & 29.75 \\ \hline
\multirow{4}{*}{6} & $\surd $ &  &  & 60.93$\pm$\tiny{0.52} & 13.23$\pm$\tiny{2.09} & 21.74 \\
 & $\surd $ & $\surd $ &  & 67.17$\pm$\tiny{0.89} & 16.48$\pm$\tiny{1.48} & 27.48 \\
 & $\surd $ &  & $\surd $ & 73.17$\pm$\tiny{0.79} & 19.72$\pm$\tiny{2.49} & 29.45 \\
 & $\surd $ & $\surd $ & $\surd $ & 75.73$\pm$\tiny{1.55} & 21.10$\pm$\tiny{3.29}& 32.55 \\ \hline
\end{tabular}
}
\label{table4}
\end{table*}

\subsection{Effectiveness of each component}
To further analyze the efficacy of each constituent component in our method, we conducted an ablation study in Table \ref{table4}. 
From Table \ref{table4}, under varying K-shot settings, every constituent component in our method consistently enhances performance over the baseline (CoCoOp) on the FGSCR-42 dataset.
Specifically, as the number of shots increases, the effectiveness of our approach and its components markedly improves.
Additionally, we respectively incorporated the multi-class text prompt and the remote-sensing bias in both text and visual prompts.
These additions, even under varied shot conditions in the open FGSCR-42 dataset, led to performance H increases of 0.7-10.81 over the baseline.

In conclusion, our ablation study confirms the effectiveness of each component. Collectively, they significantly enhance the accuracy of ship images across different shot conditions.

\subsection{Qualitative results}
To gain deeper insights into the operational principles of our proposed method, we present a visualization of the attention maps for both CLIP and our method in Fig. \ref{fig5}. The figure is organized into four groups, each featuring images from distinct ship categories.
In each group, the sequence from left to right displays the raw image, the attention map generated by the CLIP model, and the attention map produced by our method. A comparative analysis reveals that our method effectively focuses on specific regions of interest, such as the ship target, while the CLIP model tends to be more responsive to complex backgrounds.
Particularly, during the training phase on base classes, our method prioritizes the extraction of information over the alteration of existing knowledge by integrating detailed textual information and remote sensing priors. This approach results in a model knowledge base with minimal interference from irrelevant features.
As a result, our method effectively bridges the domain gap between base classes and new classes, leading to enhanced generalization capabilities. The quantization results serve as a testament to the superior generalization performance of our approach, providing clear and accurate explanations for its effectiveness.

\begin{table*}[!t]
\caption{
Comparison with different methods in different K-shot settings on the FGSCR-42 dataset with different backbones (ViT-B/16 and ResNet-50). H: Harmonic mean.}
\centering
\renewcommand\arraystretch{0.9}
\begin{tabular}{@{}>{\raggedright\arraybackslash}p{1.5cm}>{\raggedright\arraybackslash}p{1.2cm}
>{\raggedright\arraybackslash}p{0.8cm}cccccccc@{}}
\toprule
    \multirow{2}{*}{Backbones} & \multirow{2}{*}{Shots} & \multicolumn{3}{c}{CoOp} & \multicolumn{3}{c}{LoRA-CLIP} & \multicolumn{3}{c}{Ours} \\
    \cmidrule(r){3-5} \cmidrule(r){6-8} \cmidrule(r){9-11}
     &  & Base & New & H & Base & New & H & Base & New & H \\
    \midrule
    \multirow{2}{*}{ViT-B/16} & K=4 & 35.57 & 16.07 & 22.14 & 52.71 & 12.50 & 20.21 &67.67 & 22.22 & 33.45 \\
     & K=8 & 56.40 & 13.23 & 21.43 & 68.71 & 11.79 & 20.13 & 73.67 & 21.91 & 33.76 \\
    \bottomrule
  \end{tabular}
\label{table6}
\end{table*}
\subsection{Comparisons with LoRA-based methods}
To further validate our method's superiority, we incorporated the LoRA-based approach. Although LoRA is widely recognized in the CV field, its combination with CLIP methods is less common compared to prompt tuning methods. We integrated LoRA into the CLIP model to accomplish the base-to-new generation task, referencing recent work \cite{lora}. This work proposed a CLIP-LoRA method that dynamically inserts LoRA layers across image or text encoders, or both, and fine-tunes models by unfreezing positional embeddings, logit scales, and projections. We applied this to our base-to-new generation task in the remote sensing ship domain, with experimental results shown in Table \ref{table6}.

\begin{table*}[!t]
\caption{
Comparison with different methods in different K-shot settings on the FGSCR-42 dataset with different backbones (ViT-224*224 and VIT-334*334). H: Harmonic mean.}
\centering
\renewcommand\arraystretch{0.9}
\begin{tabular}{@{}>{\raggedright\arraybackslash}p{1.5cm}>{\raggedright\arraybackslash}p{1.2cm}
>{\raggedright\arraybackslash}p{0.8cm}cccccccc@{}}
\toprule
    \multirow{2}{*}{Backbones} & \multirow{2}{*}{Shots} & \multicolumn{3}{c}{CoOp} & \multicolumn{3}{c}{CoCoOp} & \multicolumn{3}{c}{Ours} \\
    \cmidrule(r){3-5} \cmidrule(r){6-8} \cmidrule(r){9-11}
     &  & Base & New & H & Base & New & H & Base & New & H \\
    \midrule
    \multirow{2}{*}{ViT-224} & K=4 & 35.57 & 16.07 & 22.14 & 42.11 & 16.50 & 23.71 &67.67 & 22.22 & 33.45 \\
     & K=8 & 56.40 & 13.23 & 21.43 & 60.93 & 13.23 & 21.74 & 73.67 & 21.91 & 33.76 \\
    \midrule 
    \multirow{2}{*}{ViT-334} & K=4 & 36.78 & 17.28 & 23.51 & 45.70 & 17.29 & 25.09 & 68.71 & 24.16 & 35.75 \\
     & K=8 & 58.78 & 15.30 & 24.28 & 62.39 & 16.78 & 26.45 & 76.42 & 22.69 & 34.09 \\
    \bottomrule
  \end{tabular}
\label{table7}
\end{table*}
Table \ref{table6} shows that LoRA-based methods exhibit varying performance in the base-to-new generation task. They fit well on base classes but show poor generalization on new classes. This aligns with our understanding because, in the remote sensing ship domain, sample sizes are very small and sample complexity is high. The LoRA method, with learnable parts in both image and text encoders, tends to overfit base class knowledge, leading to poor generalization. Therefore, we believe prompt tuning methods are more suitable for the base-to-new generation task in the ship domain.

\subsection{Performances on larger images}
In real-world scenarios, remote sensing images require high-resolution applications. Our current CLIP prompt tuning method is based on a 224x224 resolution. To better meet practical needs, we verified our method's effectiveness at higher resolutions. Using CLIP's pre-trained 336x336 resolution model, we incorporated various prompt tuning methods and evaluated their effectiveness. The results are shown in Table \ref{table7}.

Table \ref{table7} clearly demonstrates that higher resolutions improve performance for the base-to-new generation task in the remote sensing ship domain, enhancing both base and new classes. This improvement is intuitive because remote sensing images have significant redundancy, with effective information such as classification targets occupying a small portion of the image. Thus, increasing the resolution significantly enhances information extraction.

\section{Conclusion}
In this study, we introduce a novel prompt tuning method, drawing inspiration from VLMs, specifically designed to tackle the base-to-new generalization challenge in RS-FGSC. Our approach refines domain-specific prompts with a hierarchical, multi-granularity design and integrates remote sensing ship priors via bias terms learned from small, trainable networks. This strategy enables the learning of a generalizable feature representation.
Additionally, we contribute the FGSCM-52 dataset, a comprehensive public resource comprising 52 classes for RS-FGSC, which addresses the need for a more extensive and diverse dataset in this domain.
Our extensive experimental evaluation of benchmark datasets confirms the superior effectiveness of our method in the RS-FGSC field. The results underscore the potential of our approach to advance the state-of-the-art in fine-grained ship classification and to inspire future research in this dynamic field.

\section{Acknowledgment}
This work was supported by the National Natural Science Foundation of China (No. 62376282).
\bibliographystyle{IEEEtran}
\bibliography{reference}

\end{document}